\documentclass[letterpaper, 10 pt, conference]{ieeeconf}  

\IEEEoverridecommandlockouts                              

\usepackage[utf8]{inputenc}
\usepackage{graphics} 
\usepackage{graphicx}
\usepackage{amsmath} 
\usepackage{amssymb}  
\usepackage{hyperref}
\usepackage{xcolor}
\usepackage{siunitx}
\usepackage[backend=biber,style=ieee,natbib=true, maxcitenames=1,mincitenames=1,maxbibnames=5]{biblatex}
\usepackage{footmisc}
\usepackage[capitalise]{cleveref}
\usepackage{multirow}
\usepackage{amsmath}
\usepackage{comment}
\usepackage{makecell}
\usepackage{gensymb}

\let\labelindent\relax
\usepackage{enumitem}
\setlist[itemize]{leftmargin=*}
\setlist[enumerate]{leftmargin=*}

\graphicspath{ {images/} }

\newcommand{\norm}[1]{\left\lVert#1\right\rVert}

\renewcommand{\vec}[1]{\mathbf{#1}}

\newcommand{\R}{\mathbb{R}}
\newcommand{\N}{\mathbb{N}}

\newcommand{\va}{\mathbf{a}}
\newcommand{\ve}{\mathbf{e}}
\newcommand{\vf}{\mathbf{f}}
\newcommand{\vF}{\mathbf{F}}
\newcommand{\vg}{\mathbf{g}}
\newcommand{\vx}{\mathbf{x}}
\newcommand{\vu}{\mathbf{u}}
\newcommand{\vomega}{\boldsymbol \omega}
\newcommand{\vtau}{\boldsymbol \tau}

\DeclareMathOperator*{\minimize}{\text{minimize}}
\DeclareMathOperator{\Tr}{Tr}

\title{\LARGE \bf
Sim-to-(Multi)-Real: Transfer of Low-Level Robust Control Policies to Multiple Quadrotors 
}

\author{Artem Molchanov$^{*}$, Tao Chen$^{*}$, Wolfgang H\"onig, James A. Preiss, Nora Ayanian and Gaurav S. Sukhatme
\thanks{*Equal contribution}
\thanks{All authors are with the Computer Science department, University of Southern California, Los Angeles, California, USA
        {\tt\small \{molchano, taochen, whoenig, japreiss, ayanian, gaurav\}@usc.edu}}%
}

\begin{document}

\maketitle
\thispagestyle{empty}
\pagestyle{empty}

\begin{abstract}
Quadrotor stabilizing controllers often require careful, model-specific tuning for safe operation.
We use reinforcement learning to train policies in simulation that transfer remarkably well to multiple different physical quadrotors.
Our policies are low-level, i.e., we map the rotorcrafts' state directly to the motor outputs.
The trained control policies are very robust to external disturbances and can withstand harsh initial conditions such as throws.
We show how different training methodologies (change of the cost function, modeling of noise, use of domain randomization) might affect flight performance.
To the best of our knowledge, this is the first work that demonstrates that a simple neural network can learn a robust stabilizing low-level quadrotor controller (without the use of a stabilizing PD controller) that is shown to generalize to multiple quadrotors.
The video of our experiments can be found at \texttt{\url{https://sites.google.com/view/sim-to-multi-quad}}.
\end{abstract}

\section{INTRODUCTION}
Traditional control-theoretic approaches to stabilizing a quadrotor often require careful, model-specific system identification and parameter tuning to succeed.
We are interested in finding a single control policy that stabilizes any quadrotor and moves it to a goal position safely, without manual parameter tuning.
Such a control policy can be very useful for testing of new custom-built quadrotors, and as a backup safety controller.
Our primary objective for the controller is robustness to external disturbances and recovery from collisions.
Our secondary objective is position control and trajectory tracking.
In our work, we use reinforcement learning trained on simulated quadrotor models to learn a control policy.
To close the gap between simulation and reality, we analyze the impact of various key parameters, such as modeling of sensor noise and using different cost functions.
We also investigate how we can improve sim-to-real transfer (S2R) by applying domain randomization, a technique that trains over a distribution of the system dynamics to help trained policies be more resistant to a simulator's discrepancies from reality~\cite{tobin17iros, sadeghi17rss,JamesCoRL2017,tan18rss}.
We transfer our policy to three different quadrotor platforms and provide data on hover quality, trajectory tracking, and robustness to strong disturbances and harsh initial conditions.

To the best of our knowledge, ours is the first neural network (NN) based low-level quadrotor attitude-stabilizing (and trajectory tracking) policy trained completely in simulation that is shown to generalize to multiple quadrotors. 
Our contributions can be summarized as follows:
\begin{itemize}
    \item A system for training model-free low-level quadrotor stabilization policies without any auxiliary pre-tuned controllers.
    \item Successful training and transfer of a single control policy from simulation to multiple real quadrotors.
    \item An investigation of important model parameters and the role of domain randomization for transferability.
    \item A software framework for flying Crazyflie 2.0 (CF) based platforms using NN controllers and a Python-based simulated environment compatible with OpenAI Gym~\cite{gym} for training transferable simulated policies\footnote{\label{note:pubtime}Will be available by the date of the official publication.}.
\end{itemize}

\begin{figure}
    \includegraphics[width=\columnwidth]{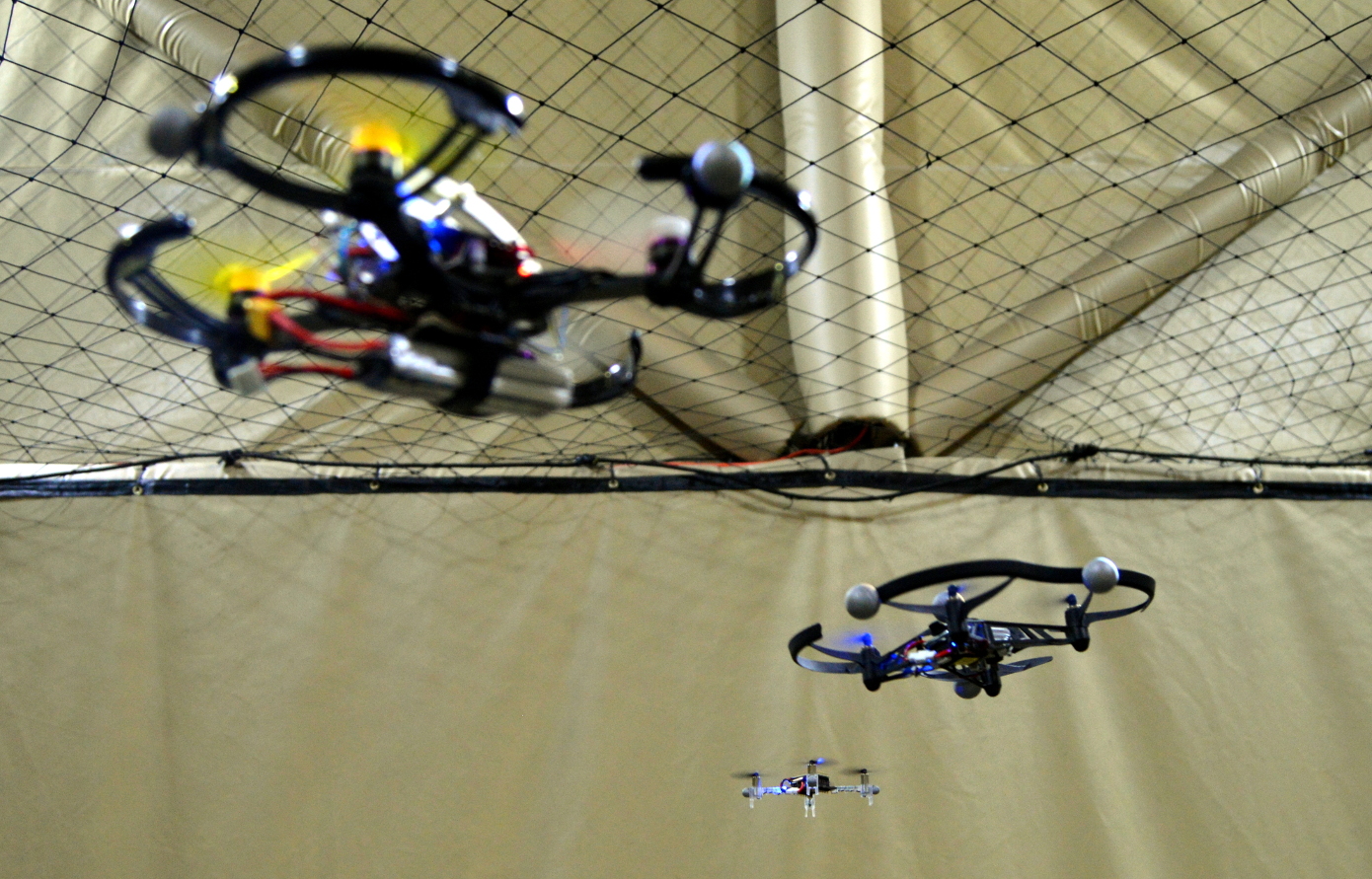}
    \caption{Three quadrotors of different sizes controlled by the same policy trained entirely in simulation.}
    \label{fig:3_quads}
    \vspace{-5mm}
\end{figure}

\section{RELATED WORK}
Transferring from simulation to reality (S2R) is a very attractive approach to overcome the issues of safety and complexity of data collection for reinforcement learning on robotic systems.
In the following we group the related work into different categories.

\textbf{S2R with model parameter estimation.}
A substantial body of work considers closing the S2R gap by carefully estimating parameters of the real system to achieve a more realistic simulation.
For example, \citet{lowrey18simpar} transfer a non-prehensile manipulation policy for a system of three one-finger robot arms pushing a single cylinder. 
\citet{tan18rss} show transferability of agile locomotion gaits for quadruped robots. \citet{antonova17arxiv} learn a robust policy for rotating an object to a desired angle.
While careful parameter estimation can provide a good estimate of the model, it often requires sophisticated setups~\cite{forster15}. 
To avoid these complications, we focus on transferring to novel quadrotors for which we do not perform accurate model parameter estimation.

\textbf{S2R with iterative data collection.}
An alternative way to overcome the problem of S2R gap is learning distributions of dynamics parameters in an iterative manner.
For example, \citet{christiano16arxiv} learn inverse dynamics models from data gradually collected from a real robotics system, while transferring trajectory planning policy from a simulator.
\citet{chebotar19icra, zhu18ijcai} transfer manipulation policies by iteratively collecting data on the real system and updating a distribution of dynamics parameters for the simulator physics engine.
Similar principles work for the problem of humanoid balancing~\cite{tan16iros}.

The common problem of these approaches is the necessity to execute untrained policies directly on the robot, which may raise safety concerns. 
In contrast to all of the works presented above, we are interested in a method that can 
i) transfer very low-level policies that directly control actuator forces, 
ii) transfer to multiple real robots with different dynamics, 
iii) control inherently unstable systems with dangerous consequences of failures, 
iv) avoid the need for data collection on the real system with a possibly unstable policy, and 
v) avoid complex setups for system parameter estimation.

\textbf{Domain randomization.}
Domain randomization~(DR)~\cite{tobin17iros} is a simple albeit promising domain adaptation technique that is well suited for S2R. 
It compensates for the discrepancy between different domains by extensively randomizing parameters of the training (source) domain in simulation.
In some cases, it can eliminate the need for data collection on the real robot completely.
DR has been successfully applied for transferring visual features and high level policies.
\citet{tobin17iros} employ DR for training a visual object position predictor for the task of object grasping.
The policy is trained in a simulation with random textures and lightning. 
A similar direction explores intermediate-level representations, such as object corners identification, and trained an interpretable high-level planning policy to stack objects using the Baxter robot~\cite{tremblay18}.

S2R transfer of dynamics and low-level control policies is considered a more challenging task due to the complexity of realistic physics modeling.
Nonetheless, there have been some promising works.
\citet{peng18icra} apply DR to the task of pushing an object to a target location using a Fetch arm.
The policy operates on joint angle positions instead of directly on torques.
\citet{mordatch15iros} transfer a walking policy by optimizing a trajectory on a small ensemble of dynamics models.
The trajectory optimization is done offline for a single Darwin robot.
In contrast, our work does not require careful selection of model perturbations and is evaluated on multiple robots.

\textbf{S2R for quadrotor control.}
S2R has also been applied to quadrotor control for transferring high-level visual navigation policies.
Most of these works assume the presence of a low-level controller capable of executing high-level commands.
\citet{sadeghi17rss} apply CNN trained in simulation to generate a high level controller selecting a direction in the image space that is later executed by a hand-tuned controller. 
\citet{kang19icra} look at the problem of visual navigation using a Crazyflie quadrotor and learn a high-level yaw control policy by combining simulated and real data in various ways.

The approaches most related to ours are the works of \citet{neuroflight} and \citet{hwangbo17iros}.
In the former work, a low-level attitude controller is replaced by a neural network and transferred to a real quadrotor~\cite{neuroflight}.
In the latter work, a low-level stabilizing policy for the Hummingbird quadrotor that is trained in simulation is transferred to the Hummingbird.
In contrast to their work 
i) we assume minimal prior knowledge about quadrotor's dynamics parameters; 
ii) we transfer a single policy to multiple quadrotor platforms;
iii) we simplify the cost function used for training the policy;
iv) we investigate the importance of different model parameters and the role of domain randomization for S2R transfer of quadrotor's low-level policies; and
v) unlike \citet{hwangbo17iros} we do not use auxiliary pre-tuned PD controller in the learned policy.

\section{Problem Statement}
\label{sec:problem_statement}
\begin{figure}
    \includegraphics[width=\columnwidth]{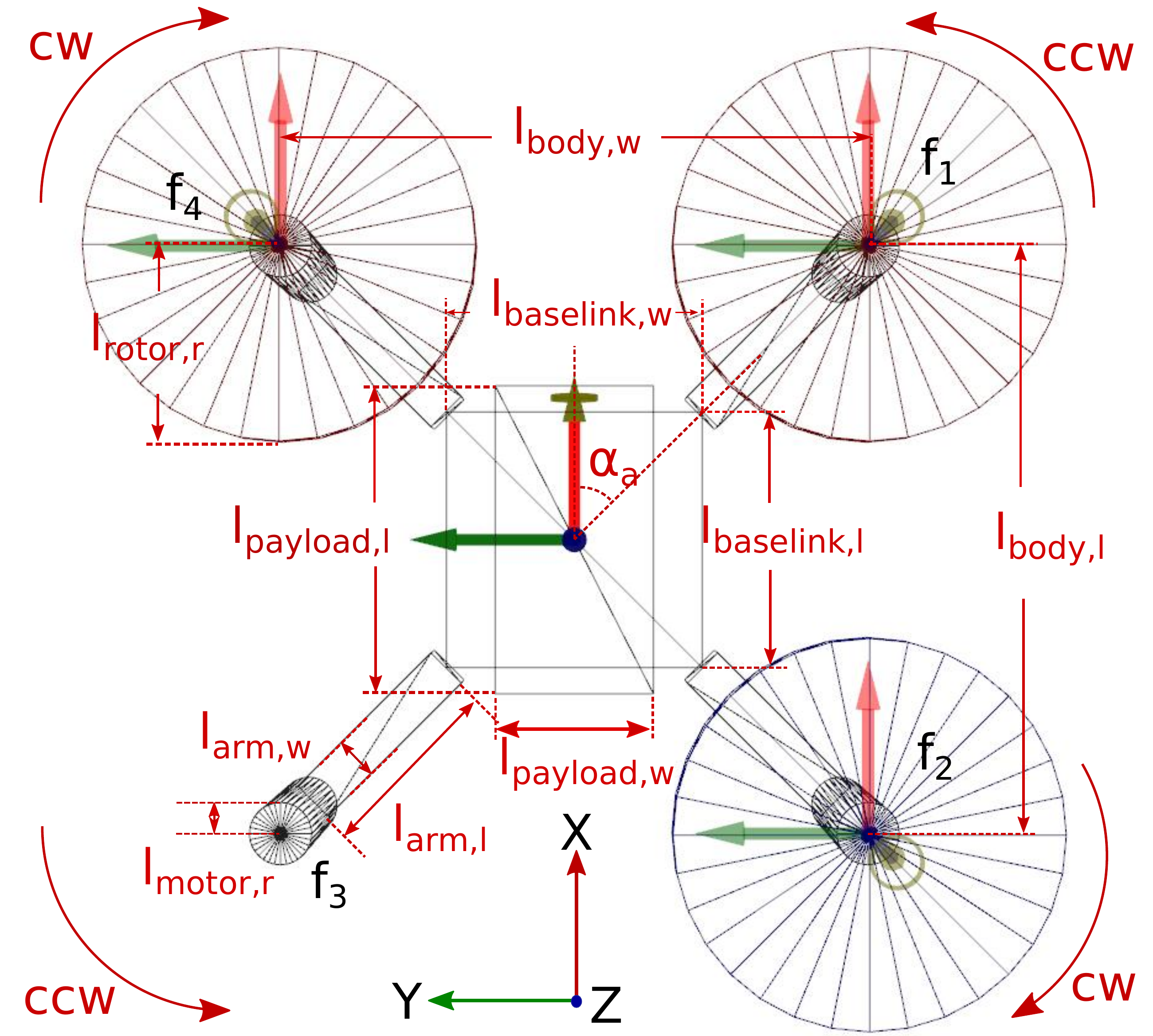}
    \caption{Top-down view of our generalized quadrotor model. 
    There are 5 components: baselink, payload, 4 arms, 4 motors, and 4 rotors. The model always assumes the $\times$ configuration, with the front pointing at the positive $X$ direction and the left pointing at the positive $Y$ direction. 
    Motors are indexed in the clockwise direction starting from the front-right motor.
    The front-right motor rotates counterclockwise and generates a thrust force $f_1$.}
    \label{fig:quad}
    \vspace{-5mm}
\end{figure}

We aim to find a policy that directly maps the current quadrotor state to rotor thrusts. The quadrotor state is described by the tuple $(\ve_p, \ve_v, R, \ve_\omega)$, where $\ve_p \in \R^3$ is the position error, $\ve_v \in \R^3$ is the linear velocity error of the quadrotor in the world frame, $\ve_\omega$ is the angular velocity error in the body frame, and $R \in SO(3)$ is the rotation matrix from the quadrotor's body coordinate frame to the world frame. The objective is to minimize the norms of $\ve_p, \ve_v, \ve_{\omega}$ and drive the last column of $R$ to $[0, 0, 1]^T$ in the shortest time.
The policy should be robust, i.e., it should be capable of recovering from different initial conditions, as well as transferable to other quadrotor platforms while retaining high performance.
We make the following assumptions:
\begin{itemize}
    \item We only consider quadrotors in the $\times$ configuration~(\cref{fig:quad}). 
    This configuration is the most widely used, since it allows convenient camera placement.
    It would not be possible for a single policy to control both $+$ and $\times$ configurations without some input specifying the configuration of the current system, since the fundamental geometric relationship between the motors and the quadrotor's axes of rotation is altered.
    \item We assume access to reasonably accurate estimates of the quadrotor's position, velocity, orientation, and angular velocity.
    Similar assumptions are typical in the robotics literature.
    They are commonly satisfied in practice by fusion of inertial measurements with localization from one or more of the following: vision, LIDAR, GPS, or an external motion capture system.
    \item We consider quadrotors within a wide but bounded range of physical parameters\footnote{This assumption is introduced to restrict our system to typical quadrotor shapes and properties. 
    We are not considering edge cases, such as very high thrust-to-weight ratios, or unusual shapes with a large offset of the center of mass with respect to quadrotor's geometric center.}.
    The ranges we experiment with are presented in \cref{tab:randomization}.
\end{itemize}

\section{Dynamics Simulation}
In this section we describe our dynamics model for simulation in detail. 

\subsection{Rigid Body Dynamics for Quadrotors}
We treat the quadrotor as a rigid body with four rotors mounted at the corners of a rectangle.
This rectangle lies parallel to the $x-y$ plane of the body coordinate frame, as do the rotors.
Each rotor is powered by a motor that only spins in one direction, hence the rotors can only produce positive thrusts in the $z$-direction of the body frame.
The origin of the body frame is placed in the center of mass of the quadrotor.
The dynamics are modeled using Newton-Euler equations:
\begin{align}
\label{eq:dynamics}
    & m \cdot \ddot{\vx} = m \cdot \vg + R \cdot \vF \\
    & \dot{\vomega} = I^{-1}(\vtau - \vomega \times (I \cdot \vomega))\\
    & \dot{R} = \vomega_{\times}R,
\end{align}
where $m \in \R_{> 0}$ is the mass, 
$\ddot{\vx} \in \R^3$ is the acceleration in the world frame, 
$\vg = [0, 0, -9.81]^T$ is the gravity vector,
$R \in SO(3)$ is the rotation matrix from the body frame to the world frame, 
$\vF \in \R^3$ is the total thrust force in the body frame, 
$\vomega \in \R^3$ is the angular velocity in the body frame,
$I \in \R^{3 \times 3}$ is the inertia tensor,
$\vtau \in \R^3$ is the total torque in the body frame, and $\vomega_{\times} \in \R^{3 \times 3}$ is a skew-symmetric matrix associated with $\vomega$ rotated to the world frame.

The total torque $\vtau$ is calculated as:
\begin{align}
    & \vtau = \vtau_{p} + \vtau_{th},
\end{align}
where 
$\vtau_{th}$ is a thruster torque, produced by thrust forces~\cite{martin10icra},
$\vtau_{p}$ is a torque along the quadrotor's $z$-axis, produced by difference in speed rotation of the propellers:
\begin{align}
    & \vtau_{p} = r_{t2t} \cdot [+1, -1, +1, -1]^T \odot \vec{f},
\end{align}
where $r_{t2t}$ is a torque to thrust coefficient, $-1$ indicates the rotor turns clockwise, $+1$ indicates the rotor turns counterclockwise, and $\vec{f}=[f_1, f_2, f_3, f_4]^T$ is a vector representing the force generated by each rotor.

\subsection{Normalized Motor Thrust Input}
It is common in the literature of quadrotor control to assume that the motor speed dynamics are nearly instantaneous.
This assumption allows us to treat each motor's thrust as a directly controlled quantity.
In this paper, to facilitate transfer of the same policy to multiple quadrotors, we instead define a normalized control input
$\hat \vf \in [0, 1]^4$
such that $\hat \vf = \mathbf{0}$ corresponds to no motor power and $\hat \vf = \mathbf{1}$ corresponds to full power.
Note that the nominal value of $\hat \vf$ for a hover state depends on the thrust-to-weight ratio of the quadrotor.
By choosing this input, we expect the policy to learn a behavior that is valid on quadrotors with different thrust-to-weight ratios without any system identification needed.
The input $\hat \vf$ is derived from the policy action $\va \in \R^4$ by the affine transformation
\begin{equation}
\label{eq:a2f}
    \hat \vf = \textstyle \frac{1}{2}(\va + \mathbf{1})
\end{equation}
to keep the policy's action distribution roughly zero-mean and unit-variance.
Since thrust is proportional to the square of rotor angular velocity, we also define
$\hat \vu = \sqrt{\hat \vf} \label{eq:f2u_}$
as the normalized angular velocity command associated with a given normalized thrust command.

\subsection{Simulation of Non-Ideal Motors}
The assumption of instantaneous motor dynamics is reasonable for the slowly-varying control inputs of a human pilot,
but it is unrealistic for the noisy and incoherent control inputs from an untrained stochastic neural network policy.
To avoid training a policy that exploits a physically implausible phenomenon of the simulator, we introduce two elements to increase realism:
motor lag simulation and a noise process.

\paragraph{Motor lag}
We simulate motor lag with a discrete-time first-order low-pass filter:
\begin{align}
    \hat \vu'_t = \frac{4 dt}{T} (\hat \vu_t - \hat \vu'_{t-1}) + \hat \vu'_{t-1},
\end{align}
where $\hat \vu'_t \in \R^4$ is the vector of the filtered normalized rotor angular velocities, $dt$ is the time between the inputs
and $T \geq 4dt$ is the 2\% \textit{settling time}, defined for a step input as
\begin{equation}
    T = dt \cdot  \min \{ t \in \N : \| \hat \vu'_{t'} - \hat \vu'_{t}\|_{\infty} < 0.02\ \text{for all}\ t' \geq t\}.
\end{equation}

\paragraph{Motor noise}
We add motor noise $\vec{\epsilon}^\vec{u}_{t}$ following a discretized Ornstein-Uhlenbeck process:
\begin{align}
    \epsilon^{\vu}_{t} = \epsilon^{\vu}_{t-1} + \theta (\mu - \epsilon^{\vu}_{t-1}) + \sigma \mathcal{N}(0, 1),
\end{align}
where $\mu = 0$ is the process mean, $\theta$ is the decay rate, $\sigma$ is the scale factor, and $\mathcal{N}(0, 1)$ is a random variable with four-dimensional spherical Gaussian distribution.

The final motor forces are computed as follows:
\begin{equation}
    \vf = f_{\max} \cdot (\hat{\vu}'_t + \epsilon^{\vu}_{t})^2.
\end{equation}

Here, $f_{max}$ is found using the \textit{thrust-to-weight ratio} $r_{t2w}$:
\begin{align}
    & f_{max} = 0.25 \cdot g \cdot m \cdot r_{t2w},
\end{align}
where $g$ is the gravity constant.

\subsection{Observation Model}
In addition to the motor noise, we also model sensor and state estimation noise.
Noise in estimation of position and orientation as well as in linear velocity is modeled as zero-mean Gaussian.
Noise in angular velocity measured by the gyroscope follows the methods presented by \citet{furrer16}.
Sensor noise parameters were roughly estimated from data recorded while quadrotors were resting on the ground.

\subsection{Numerical Methods}
We use first-order Euler's method to integrate the differential equations of the quadrotor's dynamics.
Due to numerical errors in integration, the rotational matrix $R$ loses orthogonality over time. We re-orthogonolize $R$ using the Singular Value Decomposition (SVD) method.
We perform re-orthogonolization every $0.5s$ of simulation time or when the orthogonality criteria fails:
\begin{equation}
    \|RR^T - I_{3\times3}\|_1 \geq 0.01,
\end{equation}
where $\|\cdot\|_1$ denotes the elementwise $L_1$ norm.
If $U \Sigma V^T = R$ is the singular value decomposition of $R$,
then $R_{\bot} = UV^T$ is the solution of the optimization problem
\begin{equation}
\minimize_{A \in \R^{3 \times 3}}\; \|A - R\|_F\; \text{ subject to }\; A^TA = I,
\end{equation}
where $\|\cdot\|_F$ denotes the Frobenius norm~\cite{higham-matrixnearness},
making $R_{\bot}$ a superior orthogonalization of $R$ than, e.g., that produced by the Gram-Schmidt process.

\section{Learning \& Verification}
In this section, we discuss the methodology we use to train a policy, including the domain randomization of dynamics parameters, the reinforcement learning algorithm, the policy class, and the basic structure of our experimental validations.
More details of the experiments are given in~\cref{sec:experiments}.

\subsection{Randomization}

\begin{table}[t]
\centering
\caption{Randomization variables and their distributions.}
\begin{tabular}{c|c|m{1.5cm}|m{2.5cm}}
    Variable & Unit & Nominal Randomization & Total Randomization \\
    \hline
    \hline
    $m$ & \si{kg} & 0.028 & $\leq5$ \\ 
    \hline
    $l_{body,w}$ & \si{m} & 0.065 & $\sim \mathcal{U}(0.05, 0.2)$ \\ 
    \hline
    $T$ & \si{s} & 0.15 & $\sim \mathcal{U}(0.1, 0.2)$ \\
    \hline
    $r_{t2w}$ & \si{kg/N} & 1.9 & $\sim \mathcal{U}(1.8, 2.5)$ \\
    \hline 
    $r_{t2t}$ & \si{s^{-2}} & 0.006 & $\sim \mathcal{U}(0.005, 0.02)$ \\ 
\end{tabular}
\label{tab:randomization}
\end{table}

We investigate the role of domain randomization for generalization toward models with unknown dynamics parameters. 
The geometry of the generalized quadrotor is defined by variables shown in \cref{fig:quad}. 
For brevity, we omit the height variables ($l_{*,h}$, where $*$ is the component name) in the figure. \cref{tab:randomization} shows the list of variables we randomize. 

During training, we sample dynamics parameters for each individual trajectory.
We experiment with two approaches for dynamics sampling:
\begin{enumerate}
    \item Randomization of parameters around a set of nominal values assuming that approximate estimates of the parameters are available. 
    We use existing Crazyflie 2.0 parameter estimates~\cite{forster15}. 
    \item Randomization of parameters within a set of limits.
    The method assumes that the values of parameters are unknown but bound by the limits.
\end{enumerate}

In the first scenario, we randomize all parameters describing our quadrotor around a set of nominal values, and in case a Gaussian distribution is used, we check the validity of the randomized values (mostly to prevent negative values of inherently positive parameters).
In the second scenario, we start by sampling the overall width of the quadrotor ($l_{body,w}$) and the rest of the geometric parameters are sampled with respect to it.
The total mass $m$ of the quadrotor is computed by sampling densities of individual components.
The inertia tensors of individual components with respect to the body frame are found using the parallel axis theorem.

\subsection{Policy Representation}
We use a fully-connected neural network to represent a policy.
The neural network has two hidden layers with 64 neurons each and the $\tanh$ activation function, except for the output layer that has the linear activation.
The network input is an 18-dimensional vector representing the quadrotor state presented in \cref{sec:problem_statement}. 
Rather than inputting the current state and a goal state, we input only the error between the current and goal state, except for the rotation matrix which represents the current orientation.
This reduces the input dimensionality, and trajectory tracking is still possible by shifting the goal state.
In our policy we do not use any additional stabilizing PID controllers and directly control the motor thrusts, in contrast to existing approaches~\cite{hwangbo17iros}.
Hence, our neural network policy directly outputs the normalized thrust commands $\va$ that are later converted to the normalized force commands $\hat{\vf}$ (see \cref{eq:a2f}).

\subsection{Policy Learning}

The policy is trained using the Proximal Policy Optimization (PPO) algorithm~\cite{ppo}.
PPO has recently gained popularity for its robustness and simplicity.
PPO is well-suited for RL problems with continuous state and action spaces where interacting with the environment is not considered expensive.
We use the implementation available in \emph{Garage}\footnote{\href{https://github.com/rlworkgroup/garage/commit/77714c38d5b575a5cfd6d1e42f0a045eebbe3484}{https://github.com/rlworkgroup/garage}}, a TensorFlow-based open-source reinforcement learning framework.
This framework is an actively supported and growing reincarnation of the currently unsupported rllab framework~\cite{duan16icml}.

During training, we sample initial states uniformly from the following sets: orientation is sampled from the full SO(3) group, position within a \SI{2}{m} box around the goal location, velocity with a maximum magnitude of \SI{1}{m/s}, and angular velocity with a maximum magnitude of $\SI{2\pi}{rad/s}$.
The goal state is always selected to hover at $[0,0,2]^T$ in the world coordinates.
At execution time, we can translate the coordinate system to use the policy as a trajectory tracking controller. 
We parameterize the quadrotor's attitude with a rotation matrix instead of a quaternion because the unit quaternions double-cover the rotation group $SO(3)$, meaning that a policy with a quaternion input must learn that the quaternions $q$ and $-q$ represent the same rotation.

The reinforcement learning cost function  is defined as
\begin{equation}
\label{eq:cost}
\begin{split}
    c_t & = (\norm{\ve_p}_2 + \alpha_v \norm{\ve_v}_2 + \alpha_\omega \norm{\ve_\omega}_2 + \alpha_a \norm{\va}_2 \\
    & + \alpha_R \cos^{-1} \left((\Tr(R) - 1)/2\right)) dt,
\end{split}
\end{equation}
where $R$ is the rotation matrix and $\alpha_\omega, \alpha_a, \alpha_R, \alpha_v$ are non-negative scalar weights.
The term $\cos^{-1} \left((\Tr(R) - 1)/2\right)$ represents the angle of rotation between the current orientation and the identity rotation matrix.
We investigate influence of different components in the experimental section.

\subsection{Sim-to-Sim Verification}
Before running the policy on real quadrotors, the policy is tested in a different simulator.
This sim-to-sim transfer helps us verify the physics of our own simulator and the performance of policies in a more realistic environment. 
In particular, we transfer to the Gazebo simulator with the RotorS package~\cite{furrer16} that has a higher-fidelity simulation compared to the one we use for training.

Gazebo by default uses the ODE physics engine, rather than our implementation of the Newton-Euler equations.
{RotorS} models rotor dynamics with more details, e.g. it models drag forces which we neglect during learning.
It also comes with various pre-defined quadrotor models, which we can use to test the performance of trained policies for quadrotors where no physical counterpart is available.

We found that using our own dynamics simulation for learning is faster and more flexible compared to using Gazebo with RotorS directly.

\subsection{Sim-to-Real Verification}%
\label{sec:approach:sim_to_real}

We verify our approach on various physical quadrotors that are based on the Crazyflie 2.0 platform.
The Crazyflie 2.0 is a small quadrotor that can be safely operated near humans. Its light weight (\SI{27}{g}) makes it relatively crash-tolerant.
The platform is available commercially off-the-shelf with an open-source firmware.
We build heavier quadrotors by buying standard parts (e.g., frames, motors) and using the Crazyflie's main board as a flight controller.

We test policies by sequentially increasing quadrotor size (starting with the Crazyflie 2.0) for safety reasons.
We quantify the performance of our policies using three different experiments.
First, we evaluate the hover quality by tasking the quadrotor to hover at a fixed position and record its pose at a fixed frequency.
For each sample, we compute the Euclidean position error $\norm{\ve_p}$ and the angular error ignoring yaw:
\begin{equation}
    e_\theta = \arccos (R(:,3) \cdot [0, 0, 1]^T) = \arccos R(3,3),
\end{equation}
where $R(:,3)$ is the last column of the rotation matrix $R$, and $R(3,3)$ is its bottom-right element.
We denote the mean of the position and angular errors over all collected hover samples as $e_h$ (in \si{m}) and $\bar e_\theta$ (in \si{deg}), respectively.
We characterize oscillations by executing a fast Fourier transform (FFT) on the roll and pitch angles, and report $f_o$ (in \si{Hz}) -- the highest frequency with a significant spike.
Second, we evaluate the trajectory tracking capabilities by tasking the quadrotor to track a pre-defined figure-eight trajectory and record the position errors $\norm{\ve_p}$ (in \si{m}).
We denote the mean of the errors during the flight as $e_t$.
Finally, we disturb the quadrotors and check if they recover using our policies (an experiment that is difficult to quantify on a physical platform.)

\section{Experiments}
\label{sec:experiments}
\begin{table}[t]
\centering
\caption{Robot Properties.
}
\begin{tabular}{l||c|c|c}
    Robot & CF & Small & Medium\\
    \hline
    \hline
    Weight [\si{g}] & 33 & 73 & 124\\
    $l_{body,w}$ [\si{mm}] & 65 & 85 & 90\\
    $l_{rotor,r}$ [\si{mm}] & 22 & 33 & 35\\
    $r_{t2w}$ (approximate) & 1.9 & 2.0& 2.7
\end{tabular}
\label{tab:robots}
\vspace{-5mm}
\end{table}

We validate our control policies on three different quadrotors with varying physical properties: Crazyflie 2.0, small, and medium size as described in \cref{tab:robots}.
All quadrotors use a similar control board with a STM32F405 microcontroller clocked at \SI{168}{Mhz}, executing the same firmware.
We use the Crazyswarm testbed~\cite{crazyswarm} for our experiments.
In particular, the state estimate is computed by an extended Kalman filter (EKF) that fuses on-board IMU data and motion capture information.
For the experiments with trajectories, we upload them at the beginning of the flight and compute the moving goal states on-board.
We make three major changes to the firmware:
First, we add a control policy, which is an auto-generated C-function from the trained NN model in TensorFlow. 
Second, we remove the software low-pass filter of the gyroscope, and increase the bandwidth of its hardware low-pass filter.
We found that the reduction in the gyroscope delay significantly reduces the quadrotor's physical oscillations when controlled by our policy.
Third, we only use the motion capture system to estimate linear velocities using finite differences and ignore accelerometer readings.
We found that the velocity estimates were noisier otherwise, which caused a large position offset when using our policy.
Whenever we compare to a non-learned controller, we use the default Crazyswarm firmware without our modifications.
Our motion capture system captures pose information at \SI{100}{Hz}; all on-board computation (trajectory evaluation, EKF, control) is done at \SI{500}{Hz}.
Evaluating the neural network takes about \SI{0.8}{ms}.

To train the policy, we collect $40$ simulated trajectories with a duration of \SI{7}{s} (i.e.~\SI{4.7}{min} of simulated flight) at each iteration of PPO.
In simulation the policy runs at \SI{100}{Hz} and the dynamics integration is executed at \SI{200}{Hz}.
Samples for training are collected with the policy rate.
We train the majority of our policies for $3000$ iterations, which we found sufficient for convergence.
The exception is the scenarios with randomization, for which we run $6000$ iterations due to slower convergence.
In each scenario, we train five policies by varying the seed of the pseudorandom number generator used to generate the policy's stochastic actions and inject noise into the environment.
For the test on the real system, we select the two best seeds according to the average (among trajectories) sum-over-trajectory Euclidean distance cost (i.e. $\norm{\ve_p}$) computed during policy training.
After that, we visually inspect the performance of the two seeds in simulation and select the one that generates smoother trajectories and exhibits smaller attitude oscillations (a subjective measure).

\subsection{Ablation Analysis on Cost Components}
\label{sec:exp:ablation}

We analyze the necessity of different terms in the RL training cost function~\eqref{eq:cost} on the flight performance in simulation and on a real quadrotor, because we are interested in a simpler cost function with fewer hyper-parameters.
During training, we use approximate parameters of the Crazyflie 2.0 quadrotor model~\cite{forster15}.
Here, we do not apply parameter randomization, but we incorporate sensor and thrust noise.
We let the quadrotor hover at a fixed point and record its pose at \SI{100}{Hz} for \SI{10}{s} and report the mean position error $e_h$, mean angular error $\bar \ve_\theta$, and oscillation frequency $f_o$, as defined in \cref{sec:approach:sim_to_real}. 
Our results are shown in \cref{tab:ablation}.

We notice that we can train a successful policy with a cost that only penalizes position, angular velocity, and actions, as long as $\alpha_\omega$ is larger than $0.05$ but smaller than 1 (see rows 1 -- 6 in \cref{tab:ablation}).
The optimal value of $\alpha_\omega$ differs slightly: in simulation $\alpha_\omega=0.25$ achieves the lowest position and angular errors.
On the real quadrotor, we notice that higher $\alpha_\omega$ can result in significantly higher errors in position.
Thus, we chose $\alpha_\omega=0.1, \alpha_a=0.05$ (and $\alpha_R=\alpha_v=0$) as a baseline for our further experiments.

We can add a cost for rotational errors by setting $\alpha_R > 0$, which improves position and angular errors in simulation, but results in slightly larger position errors on the physical quadrotor (see rows 7 and 8 in \cref{tab:ablation}) compared to the baseline.
The major advantage of this added cost is that it also stabilizes yaw, which might be desired for takeoff or if the quadrotor is carrying a camera.

Finally, we compared our cost function with the cost function that is similar to the one previously introduced by~\cite{hwangbo17iros}. 
It additionally includes cost on linear velocity (i.e. $\alpha_v > 0$; see row 9 in \cref{tab:ablation}).
This cost function is harder to tune because of the larger number of hyper-parameters.
The learned policy showed slightly worse position and angular errors in simulation and on the physical robot.
All policies did not show any significant oscillations ($f_o\leq \SI{1.2}{Hz}$).

\begin{table}
\centering
\caption{Ablation Analysis on Cost Components.\hspace{\textwidth}
$\alpha$ values not listed are 0.}
\setlength{\tabcolsep}{3pt}
\begin{tabular}{c|p{3cm}||c|c|c||c|c|c}
 \# & Cost & \multicolumn{3}{c||}{RotorS} & \multicolumn{3}{c}{CF} \\
 & Parameters & $e_h$ & $\bar e_\theta$ & $f_o$ & $e_h$ & $\bar e_\theta$ & $f_o$\\
 \hline
 \hline
 1 & $\alpha_\omega:0.00$, $\alpha_a:0.05$    & \multicolumn{6}{c}{Training failed}\\
 2 & $\alpha_\omega:0.05$, $\alpha_a:0.05$ & \multicolumn{3}{c||}{No Takeoff} & $0.14$ & $3.52$ & $1.0$\\
 3 & $\alpha_\omega:0.10$, $\alpha_a:0.05$  & 0.05 & 0.84 & 1.0 & \bf{0.09} & $2.07$ & $0.9$\\
 4 & $\alpha_\omega:0.25$, $\alpha_a:0.05$  & 0.05 & 0.02 & 0.7 & 0.21 & 2.59 & 0.6\\
 5 & $\alpha_\omega:0.50$, $\alpha_a:0.05$  & 0.08 & 0.07 & 0.5 & 0.30 & 2.34 & 0.5\\
 6 & $\alpha_\omega:1.00$, $\alpha_a:0.05$    & \multicolumn{6}{c}{Training failed}\\
 \hline
 7 & \makecell[l]{$\alpha_\omega:0.10$, $\alpha_a:0.05$\\ $\alpha_R:0.25$} & 0.06 & 0.02 & 1.1 & 0.14 & 1.67 & 1.0\\
 8 & \makecell[l]{$\alpha_\omega:0.10$, $\alpha_a:0.05$\\ $\alpha_R:0.50$} & \bf{0.04} & \bf{0.01} & 0.8 & 0.14 & \bf{1.51} & 0.8\\
 \hline
 9 & $\alpha_\omega:0.075$, $\alpha_a:0.050$, $\alpha_R:0.000$, $\alpha_v:0.125$ (cmp. \cite{hwangbo17iros}) & \bf{0.04} & 4.31 & 1.2 & 0.14 & 3.73 & 1.0
\end{tabular}
\label{tab:ablation}
\end{table}

\subsection{Sim-to-Real: Learning with Estimated Model}
\label{sec:exp:sim_to_real}

\begin{figure}
    \includegraphics[width=\columnwidth]{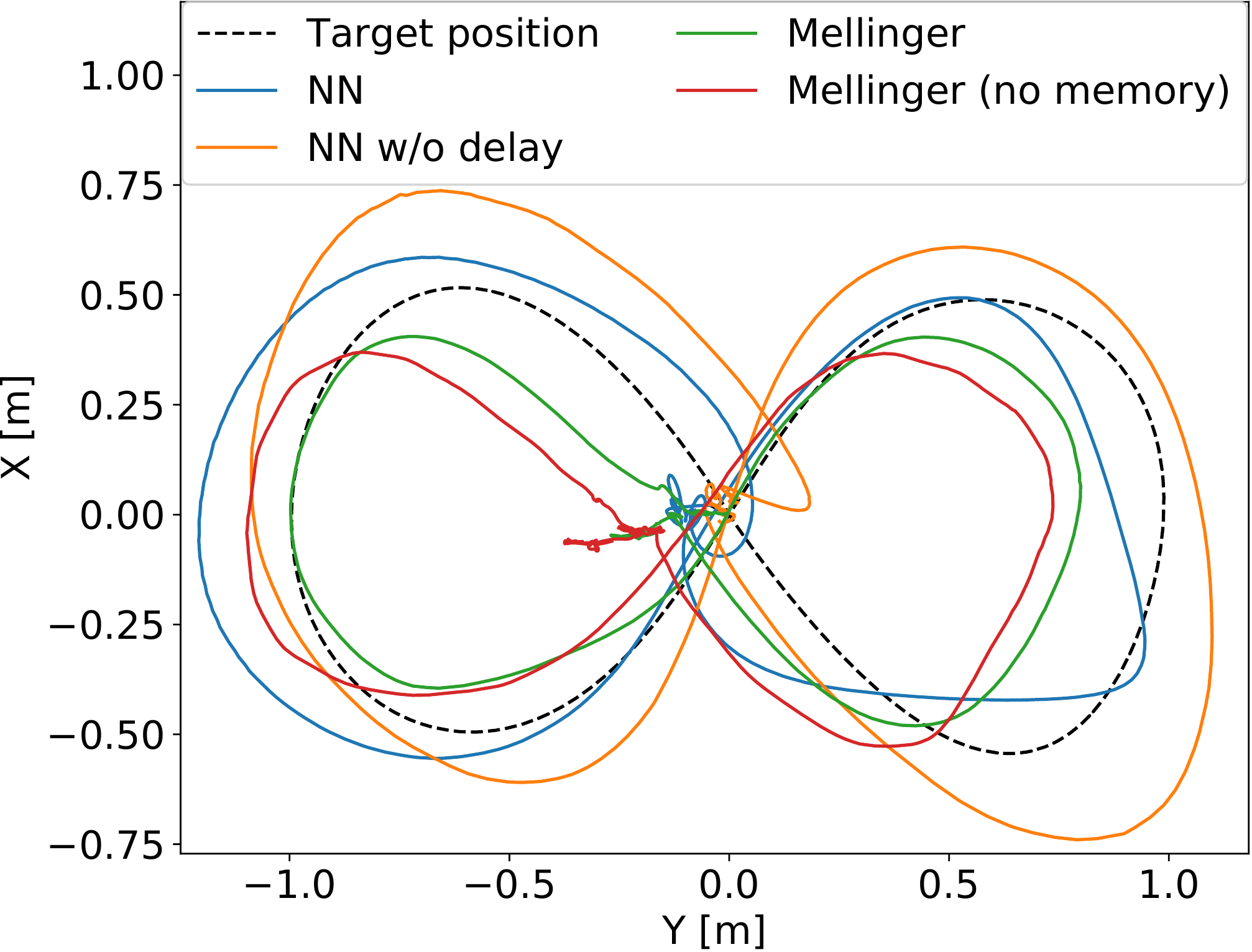}
    \caption{Trajectory tracking performance of a Crazyflie 2.0 using different controllers.
    The target trajectory is a figure-eight (to be executed in $\SI{5.5}{s}$).
    }
    \label{fig:sim_to_real_fig8}
    \vspace{-5mm}
\end{figure}

\begin{table*}
\centering
\caption{Sim-to-Multi-Real Results.}
\begin{tabular}{c|c||c|c|c||c|c|c||c|c|c}
\# & Policy & \multicolumn{3}{c||}{CF} & \multicolumn{3}{c||}{Small} & \multicolumn{3}{c}{Medium}\\
       & & $\bar e_\theta$ & $f_o$ & $e_t$ & $\bar e_\theta$ & $f_o$ & $e_t$ & $\bar e_\theta$ & $f_o$ & $e_t$\\
 \hline
 \hline
1 & Mellinger & 1.93 & 1.4 & 0.11 & 1.11 & 0.5 & 0.14 & 0.78 & 1.1 & 0.04\\
2 & Mellinger (no memory) & 3.68 & 6.2 & 0.20 & 1.23 & 0.8 & 0.16 & 5.53 & 5.5 & 0.07\\
3 & Mellinger (uniform) & 3.17 & 6.3 & 0.19 & 1.12 & 0.7 & 0.19 & 5.29 & 5.4 & 0.32\\
\hline
4 & NN CF & 1.53 & 0.9 & 0.19 & 1.13 & 1.0 & 0.21 & 2.99 & 1.0 & 0.47\\
5 & NN CF (w/o delay) & 1.90 & 1.0 & 0.21 & 0.93 & 0.8 & 0.21 & 2.25 & 1.0 & 0.42\\
\hline
6 & NN CF 10\% random & 1.61 & 1.1 & 0.30 & 1.34 & 1.0 & 0.22 & 3.51 & 1.0 & 0.47\\
7 & NN CF 20\% random & 1.53 & 1.1 & 0.20 & 1.12 & 1.0 & 0.21 & 1.67 & 0.9 & 0.33\\
8 & NN CF 30\% random & 2.65 & 1.4 & 0.23 & 4.33 & 1.0 & 0.24 & 1.96 & 1.0 & 0.33\\
9 & NN CF random t2w (1.5 -- 2.5) & 1.68 & 1.1 & 0.23 & 1.44 & 1.0 & 0.33 & 1.22 & 0.9 & 0.39\\
10 & NN CF random t2w (1.8 -- 2.5) & 2.32 & 1.9 & 0.21 & 1.91 & 1.0 & 0.26 & 1.70 & 1.8 & 0.49\\
\hline
11 & NN Fully random (t2w 1.5 -- 2.5) & 1.70 & 1.0 & 0.25 & 1.63 & 0.9 & 0.24 & 1.61 & 0.9 & 0.35\\
\end{tabular}
\label{tab:sim_to_multi}
\vspace{-5mm}
\end{table*}

Based on our findings in~\cref{sec:exp:ablation}, we use the cost function with parameters $\alpha_{\omega}=0.1$, $\alpha_{a}=0.05$ and test the influence of noise and motor delays (settling time) in a trajectory tracking task on the Crazyflie 2.0.
The task includes taking off, flying a figure-eight at moderate speeds (up to \SI{1.6}{m/s}, \SI{5.4}{m/s^2}, \SI{24}{deg} roll angle; \SI{5.5}{s} long), and landing.
In all cases, we do not perform model randomization.

As a baseline, we use the non-linear controller that is part of the Crazyswarm using default gains (``Mellinger''), where we observe an average Euclidean position error of \SI{0.11}{m}.
A second baseline is the same controller where we remove any computation that requires an additional state (``Mellinger (no memory)''), i.e., we set the gains for the integral terms to zero.
As expected, the average position error increases -- in this case to \SI{0.2}{m}.

Our neural network with the motor settling time $T=0.15$ has a mean position error of \SI{0.19}{m}, which is similar to the hand-tuned baseline controller without memory.
A network trained without motor delays ($T$ nearly zero) overshoots frequently and has a larger mean position error of \SI{0.21}{m}.
If the network is trained without sensor and motor noise, we measure a mean position error of \SI{0.24}{m}.
The standard deviation of the norm of the position error for the neural networks is nearly twice as high as for the non-linear feedback controller (0.06 and 0.11, respectively).
Plots for some controllers are shown in \cref{fig:sim_to_real_fig8}.

Note that none of our policies are explicitly trained for trajectory tracking.
Nonetheless, they still show competitive tracking performance compared to the baseline trajectory-tracking controller specifically tuned for the Crazyflie.

\subsection{Sim-to-Multi-Real: Learning without Model}

We now investigate how well a single policy works across different quadrotor platforms.
In all cases, we quantify the hover quality as well as trajectory tracking using the metrics defined in \cref{sec:approach:sim_to_real}.
For the medium quadrotor, we artificially limit the output RPM to $60\%$ of its maximum, to compensate for its higher thrust-to-weight ratio\footnote{Among all parameters the rough value of the thrust-to-weight ratio is relatively easy to estimate.}.
The results are shown in \cref{tab:sim_to_multi}.

We use three different baselines to provide an estimate on achievable performance. The first two baselines are identical to the Mellinger baselines in \cref{sec:exp:sim_to_real}. A third baseline is used to test transferability, in which we find a uniform set of attitude and position gains for the Mellinger controller without any memory, but keep the manually tuned values for gravity compensation in place (``Mellinger (uniform)''). This baseline provides an estimate on how well a single policy might work across different quadrotors.

We compare these ``Mellinger'' baselines to different networks, including our baseline network (BN), BN when motor delays ignored, and various policies that use randomized models during training. 
We make the following observations:
\begin{enumerate}
    \item All our policies show somewhat comparable performance to the Mellinger controllers on all platforms.
    There are no significant oscillations for the learned policies, whereas there are significant oscillations for some of the Mellinger baselines (see rows 2 and 3).
    \item Unsurprisingly, the network specifically trained for the Crazyflie works best on this platform. 
    It also performs very well on the small quadrotor, but shows large position errors on the medium quadrotor (row 4).
    Surprisingly, modeling the motor delay during training has a very small impact on tracking performance (row 5).
    \item Randomization around a set of nominal values can improve the performance, but it works best if the randomization is fairly small (20 \% in our case), see rows 6 -- 8.
    This improvement is not caused just by randomizing different thrust-to-weight ratios (rows 9 and 10).
    \item Full randomization shows more consistent results over all platforms, but performs not as well as other policies (row 11).
\end{enumerate}

\subsection{Control Policy Robustness and Recovery}

We perform recovery robustness tests by making repetitive throws of the quadrotors in the air.
We use the baseline neural network that was trained for the Crazyflie specifically on all platforms.
In these tests we do not perform comparison with the ``Mellinger'' controller since it could not recover properly from  deviations from its goal position larger than \SI{0.5}{m}.
This controller is mainly tuned for trajectory tracking with closely located points provided as full state vectors.
Our policy, on the other hand, shows a substantial level of robustness on all three platforms.
It performs especially well on the Crazyflie platform recovering from $80\%$ of all throws and up to $100\%$ throws with moderate attitude changes ($\leq35\degree$)\footnote{\label{note:percentile}Computed as a $95$-th percentile in our experiments.}.
Interestingly, it can even recover in more than half of scenarios after hitting the ground.

The policy also shows substantial level of robustness on other quadrotors. 
Similar to the Crazyflie platform, the throws with moderate attitude change do not cause serious problems to either of the platforms and they recover in $\geq 90\%$ of trials.
Stronger attitude disturbances are significantly harder and we observe roughly $50\%$ recovery rate on average.

More mild tests, like light pushes and pulling from the hover state, do not cause failures.
One observation that we make is that all policies learned some preference on yaw orientation that it tries to stabilize although we did not provide any yaw-related costs apart from the cost on angular velocities.
We hypothesize that the policy seeks a ``home'' yaw angle because it becomes unnecessary to reason about symmetry of rotation in the $xy$ plane if the quadrotor is always oriented in the same direction.

\begin{figure}
    \includegraphics[width=\columnwidth]{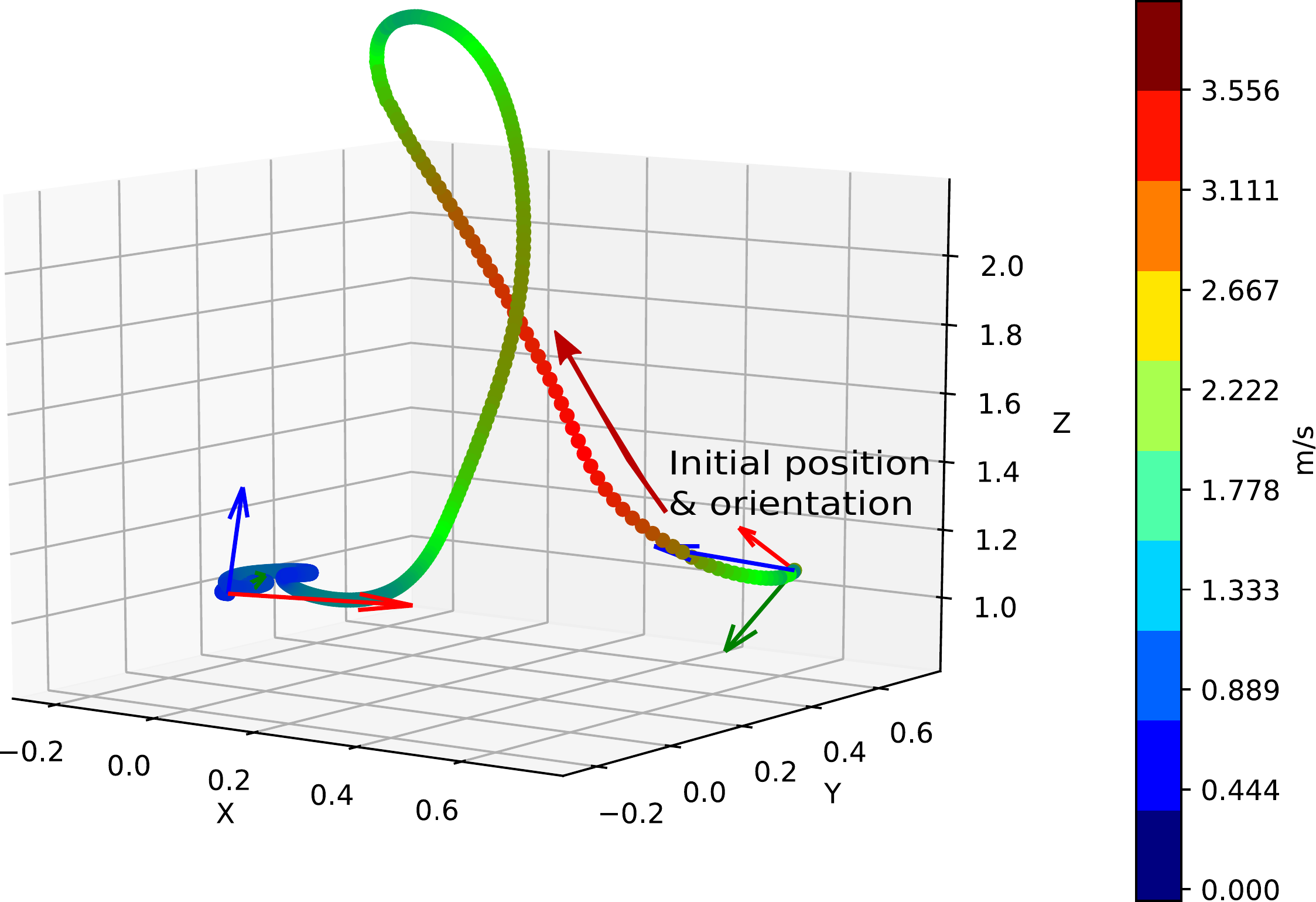}
    \caption{An example of a recovery trajectory from a random throw with an initial linear velocity of approximately \SI{4}{m/s}.}
    \label{fig:recovery}
\end{figure}

Another surprising observation is that the control policies can deal with much higher initial velocities than those encountered in training ($\leq$ \SI{1}{m/s}).
In practice, initial velocities in our tests often exceed \SI{3}{m/s} (see \cref{fig:recovery} for an example of a recovery trajectory).
The policies can take-off from the ground, thus overcoming the near-ground airflow effects.
They can also fly from distances far exceeding the boundaries of the position initialization box observed in the training. 
All these factors demonstrate strong generalization of the learned policy to the out-of-distribution states. 
Our supplemental video shows examples on the achieved robustness with all three platforms.

\section{CONCLUSIONS AND FUTURE WORK}
In this work, we demonstrate how a single neural network policy trained completely in simulation for a task of recovery from harsh initial conditions can generalize to multiple quadrotor platforms with unknown dynamics parameters.
We present a thorough study on the importance of many modeled quadrotor dynamics phenomena for the task of sim-to-real transfer.
We investigate a popular domain adaptation technique, called domain randomization, for the purpose of reducing the simulation to reality gap.

Our experiments show the following interesting results.
First, it is possible to transfer a single policy trained on a specific quadrotor to multiple real platforms, which significantly vary in sizes and inertial parameters.
Second, the transferred policy is capable of generalizing to many out-of-distribution states, including much higher initial velocities and much more distant initial positions.
Third, even policies that are trained when ignoring real physical effects (such as motor delays or sensor noise) work robustly on real systems.
Modeling such effects explicitly during training improves flight performance slightly.
Fourth, the transferred policies show high robustness to harsh initial conditions better than the hand-tuned nonlinear controller we used as a baseline.
Fifth, domain randomization is capable of improving results, but the extent of the improvement is moderate in comparison to the baseline performance trained without parameter perturbations.

Our findings open exciting directions for future work.
We are planning to explore how we can incorporate a limited number of samples collected from real systems with our policies to improve the trajectory tracking performance of our policy without manual tuning.
We are also planning to investigate if different networks are able to deal better with different thrust-to-weights ratios.
Finally, we want to explore if we can increase robustness of the policy by training with different failure cases such as broken motors.

\addtolength{\textheight}{-12cm}



\printbibliography

\end{document}